\documentclass[11pt,a4paper]{article}
\usepackage[hyperref]{emnlp2020}
\usepackage{slashbox}

\usepackage{times}
\usepackage{latexsym}

\usepackage{graphicx}

\usepackage{color, colortbl}
\definecolor{Gray}{gray}{0.9}
\definecolor{airforceblue}{rgb}{0.36, 0.54, 0.66}
\definecolor{amber}{rgb}{1.0, 0.75, 0.0}
\definecolor{applegreen}{rgb}{0.55, 0.71, 0.0}
\definecolor{blush}{rgb}{0.87, 0.36, 0.51}
\definecolor{brightube}{rgb}{0.82, 0.62, 0.91}
\definecolor{babyblue}{rgb}{0.54, 0.81, 0.94}
\definecolor{babyblueeyes}{rgb}{0.63, 0.79, 0.95}
\definecolor{bananayellow}{rgb}{1.0, 0.88, 0.21}
\definecolor{blizzardblue}{rgb}{0.67, 0.9, 0.93}
\definecolor{dandelion}{rgb}{0.94, 0.88, 0.19}
\definecolor{lavendermist}{rgb}{0.9, 0.9, 0.98}

\definecolor{blue}{rgb}{0,0,1}
\definecolor{red}{rgb}{1,0,0}
\definecolor{white}{rgb}{0,0,0}
\definecolor{blue1}{rgb}{0,0,0.1}
\definecolor{blue2}{rgb}{0,0,0.2}
\definecolor{blue3}{rgb}{0,0,0.3}
\definecolor{blue4}{rgb}{0,0,0.4}
\definecolor{blue5}{rgb}{0,0,0.5}
\definecolor{blue6}{rgb}{0,0,0.6}
\definecolor{blue7}{rgb}{0,0,0.7}
\definecolor{blue8}{rgb}{0,0,0.8}
\definecolor{blue9}{rgb}{0,0,0.9}
\definecolor{red1}{rgb}{0,0,0.1}
\definecolor{red2}{rgb}{0,0,0.2}
\definecolor{red3}{rgb}{0,0,0.3}
\definecolor{red4}{rgb}{0,0,0.4}
\definecolor{red5}{rgb}{0,0,0.5}
\definecolor{red6}{rgb}{0,0,0.6}
\definecolor{red7}{rgb}{0,0,0.7}
\definecolor{red8}{rgb}{0,0,0.8}
\definecolor{red9}{rgb}{0,0,0.9}

\usepackage{microtype}
\usepackage{multirow}
\usepackage{fixltx2e}
\usepackage{hyperref}
\usepackage{footnote}
\usepackage{booktabs}
\usepackage{algorithm,algpseudocode}

\aclfinalcopy 
\def\aclpaperid{1970} 

\title{Compositional and Lexical Semantics in RoBERTa, BERT and DistilBERT: A Case Study on CoQA}

\author{Ieva Stali\={u}nait\.{e} \and Ignacio Iacobacci \\
  Huawei Noah's Ark Lab, London, United Kingdom \\
  {\tt \{ieva.staliunaite,ignacio.iacobacci\}@huawei.com}
}

\date{}

\begin{document}
\maketitle
\begin{abstract}
Many NLP tasks have benefited from transferring knowledge from contextualized word embeddings, however the picture of what type of knowledge is transferred is incomplete. This paper studies the types of linguistic phenomena accounted for by language models in the context of a Conversational Question Answering (CoQA) task. We identify the problematic areas for the finetuned RoBERTa, BERT and DistilBERT models through systematic error analysis - basic arithmetic (counting phrases), compositional semantics (negation and Semantic Role Labeling), and lexical semantics (surprisal and antonymy). When enhanced with the relevant linguistic knowledge through multitask learning, the models improve in performance. Ensembles of the enhanced models yield a boost between 2.2 and 2.7 points in F1 score overall, and up to 42.1 points in F1 on the hardest question classes. The results show differences in ability to represent compositional and lexical information between RoBERTa, BERT and DistilBERT.
\end{abstract}

\section{Introduction}

It has recently been recognized in the research community that neural network models generally do not exploit the compositionality of language, often relying on superficial features\footnote{\url{https://2020.ieeeicassp.org/program/plenary-speakers/deep-representation-learning/}}. 
Compositionality refers to the fact that linguistic constituents combine into phrases hierarchically to compose meaning.
Contextualized word embeddings 
(BERT, \citeauthor{devlin2018bert}, \citeyear{devlin2018bert}; RoBERTa, \citeauthor{liu2019roberta}, \citeyear{devlin2018bert}; DistilBERT, \citeauthor{sanh2019distilbert},  \citeyear{sanh2019distilbert}; etc.)
can be expected to be limited in their ability to learn such complex aspects of language,
since the models are usually trained with cloze filling and next sentence prediction objectives, and are not directly exposed to semantic relations between phrases.

While larger models yield higher performance, they still lack generalization ability \cite{talmor2019olmpics} and are computationally expensive, which has led to an increasing interest in reducing model size. For instance, DistilBERT is built using the knowledge distillation technique 
\cite{bucila2006model,hinton2015distilling} 
on BERT, which leads to a lighter and faster model that does not lose much in performance on the majority of the tested tasks. On the other hand, state-of-the-art models use vast amounts of training data - 16GB for BERT, 10 times more for RoBERTa.
 
This study tackles the question of what type of linguistic knowledge is missing from contextualized word embeddings, comparing models on the basis of their training set size (BERT vs. RoBERTa) as well as their model size (BERT vs. DistilBERT). 

The tasks of machine reading comprehension (MRC) 
and dialogue are particularly fitting for this purpose, due to the fact that they require a system to interpret language within a context and perform semantic and pragmatic inference between sentences. The task in the \textbf{Co}nversational \textbf{Q}uestion \textbf{A}nswering dataset \cite[CoQA]{reddy-etal-2019-coqa} is MRC combined with dialogue - the input to the system is a context document and a dialogue of questions and answers about that text, which lead up to the question that the system is required to answer. An example from CoQA follows. 
\vspace{1mm}

\noindent \textbf{Background:} [...] At the time, the name did \underline{not} \textit{describe} a single political entity \underline{or} \textit{a distinct population of people} [...] \\
\noindent \textbf{Question n-1:} Did the name describe a political body? \\
\noindent \textbf{Answer n-1:} No \\
\noindent \textbf{Question n:} Did it \textit{describe a people group}?  \\
\noindent \textbf{Answer n:} No
\vspace{1.5mm}

In order to answer question $n$ without relying on superficial features, one needs to be able to interpret the logical operators ``\textit{and}" and ``\textit{or}" (underlined), and their scopes, as well as determine that the italicized phrases are synonymous. 
This study tackles such cases with linguistically enhanced models.
Our assumption in this paper is that if a model performs poorly on the classes of question-answer (\textsc{qa}) pairs that require certain linguistic knowledge \textsc{x} (e.g. negation, disjunction and synonymy in the example above) for their solutions, and if its performance boosts when it explicitly learns \textsc{x} (e.g. through a multitask setting with an auxiliary linguistic task), it can be considered evidence for the original model's lack of linguistic representations of \textsc{x}.

\section{Related Work}
\label{sec:relwork}

There has recently been much interest in diagnostic analysis of BERT, studying what type of linguistic representations it learns. \citet{rogers2020primer} provide an overview of such papers under the name of BERTology, showing what types of linguistic phenomena cause difficulty for BERT. 

Some such studies focus on compositional semantics in the form of negation, Negative Polarity Items (NPIs) and Semantic Role Labeling (SRL). In testing NPI licensing, \citet{warstadt2019investigating} perform a cloze task and compare whether BERT predicts a higher probability for an NPI in a licensed context or outside of such a span. They show that while BERT is capable of detecting NPI licensors (e.g. ``\textit{don't}") and NPIs themselves (e.g. ``\textit{ever}"), it only does so successfully in cases where the span of the NPI appears in the canonical position with regard to its licensor. This suggests that the model relies on word order instead of parsing the syntactic dependencies.   

When it comes to interpreting negation, \citet{ettinger2020bert} analyzes whether BERT predicts a higher probability for sentences such as ``\textit{Robin is not a tree}" and ``\textit{Robin is a bird}" than ``\textit{Robin is a tree}" and ``\textit{Robin is not a bird}", and conclude that BERT is not very sensitive to negation. 
This test, however, relies strongly on BERT's ability to represent lexical semantics of the nouns and the lack of intersection of their typical meanings. It could be argued that it therefore is not a reliable test to tell whether BERT can make logical deductions based on negation. Instead, we argue that it should be tested in a context where ``\textit{Robin}" can be anything, including a name of a tree, so that it can be determined whether BERT can infer that in such a case ``\textit{Robin}" would not be a bird.

Similarly, \citet{ettinger2020bert} has also addressed the question of whether BERT has the knowledge required to infer semantic roles from a text. For example, she tests BERT's ability to assign a higher probability to the word ``\textit{served}" in a statement such as ``\textit{the restaurant owner forgot which customer the waitress had served}" than in ``\textit{the restaurant owner forgot which waitress the customer had served}". This test is analogous to the previous example with a Robin, in the sense that it tests the model's ability to learn biases of common semantic roles that certain nouns manifest. The model can simply rely on the fact that one of these two word orders is more likely than the other, however this does not provide evidence that BERT can make inferences about semantic roles. Thus, it remains to be verified whether BERT can abstract the semantic roles from any range of naturally occurring sentences, some of which exhibit uncommon semantic role occurrences (e.g. customers serving waiters). 

In a similar vein, previous research has shown that the embeddings of antonyms in models such as BERT are not clearly distinguishable \cite{talmor2019olmpics}. This, similarly to issues with negation, shows that BERT is not good at representing non-intersecting denotations.

What is more, \citet{richardson2019probing} show that BERT performs poorly on artificially constructed diagnostic items which test the model's ability to perform logical inference. Nonetheless, they demonstrate that it is possible for BERT to extrapolate the relevant linguistic phenomena quickly by finetuning the model on the same artificial data. 

Pragmatics also plays a role in determining relations between sentences, however this field has been less explored with regard to contextualized word embeddings. Some research has probed BERT on its capabilities to infer pragmatic phenomena related to negation, such as factives, conditionals and questions \cite{jiang2019evaluating}. For instance, one has to make the pragmatic inference that a speaker would only utter ``\textit{not a one of them realized I was not human}" if their lack of humanity was already established as common knowledge. \citet{jiang2019evaluating} show that BERT takes longer to learn such complex reasoning than negation, for example. 

In contrast, some studies research what common sense knowledge and abstract reasoning that BERT and other language models learn.
\citet{talmor2019olmpics} show that there is a large gap between BERT and RoBERTa with regard to their inference abilities. For instance, since RoBERTa is trained on significantly more data, it can determine which person is older based on their ages or dates of birth, while BERT cannot. Interestingly, however, even RoBERTa is shown to rely on the range of examples seen at training time, as it is not able to generalize to the ages of people who are not born between 1920s and 2000s. This suggests that there is a need for a more abstract reasoning ability in models such as BERT, which does not seem to be solved by an increased size of the training set. Finally, \citet{ju2019technical} show that even for the RoBERTa-based abstractive model, which reached state-of-the-art results on CoQA at the time, questions with numerical answers account for a disproportionately large fraction of errors.

Based on the studies conducted so far, one general trend appears to be pertinent. That is, while many studies have explored the linguistic knowledge of BERT, it is still not clear whether BERT is able to infer compositional structures
from text as opposed to relying on biases. In addition, to the best of our knowledge, no probing tasks have been performed on BERT in the conversational question domain, 
which is fitting for analyzing BERT's behaviour in complex reasoning and inference. Finally, while larger models such as RoBERTa yield gains in performance, they still lack in generalization ability. 
Thus, this paper aims to shed light on the less scrutinized aspects of BERT's linguistic capabilities.

\section{Dataset}
\label{sec:data}

The CoQA dataset\footnote{\url{https://stanfordnlp.github.io/coqa/}} is used as a case study in this paper. It covers several domains and amounts to 127,000+ samples including a story, a \textsc{qa} pair and the dialogue history. 
The answers to the questions are based on the context document, however they can be paraphrases. The training data also contains \textit{rationales}, which are the spans of the background text containing both the answer and the context required to determine the answer. 
The test set is composed of the Reddit and Science domains, while the rest of the domains are split between train, development and test (see Table~\ref{tab:coqa}). 
Covering various domains makes CoQA diverse with regard to style and content of the dataset, whereas the addition of the dialogue history makes the dataset interesting in that it combines different language modes - a written paragraph and a conversation. Such diversity allows for a robust analysis of linguistic relations since it gives access to negation in questions as well as statements, fictional settings of unusually flipped semantic roles, counting of any abstract or concrete objects, etc. The state-of-the-art models on this dataset \cite{ju2019technical} use RoBERTa, while the dataset has not received much attention with smaller or distilled models such as DistilBERT.

\begin{table}[ht]
\begin{center}
\footnotesize{
 	\setlength{\tabcolsep}{3pt}
 	\renewcommand{\arraystretch}{1.1}
 	\centering
\begin{tabular}{lrrrr} 
\toprule 
 & \#Passages & \#\textsc{qa} & Passage & \#Turns \\
Domain & & pairs & Length & per passage \\
\midrule
\midrule
Children’s Stories      & 750   & 10.5k & 211 & 14.0 \\
Literature              & 1,815 & 25.5k & 284 & 15.6 \\
School exams   & 1,911 & 28.6k & 306 & 15.0 \\
News                    & 1,902 & 28.7k & 268 & 15.1 \\
Wikipedia               & 1,821 & 28.0k & 245 & 15.4 \\    
\midrule
Reddit                  & 100   & 1.7k  & 361 & 16.6 \\
Science                 & 100   & 1.5k  & 251 & 15.3 \\  
\midrule
Total                   & 8,399 & 127k  & 271 & 15.2 \\
\bottomrule
\end{tabular}}
\end{center}
\caption{CoQA dataset details \cite{reddy-etal-2019-coqa}.}
\label{tab:coqa}
\end{table}

\section{Baseline Models}
\label{sec:model}

The input to the model is a concatenation of the background story, the latest dialogue history of 64 tokens, and the current question. The length of the input is limited to 512 tokens. We build the baseline RoBERTa, BERT and DistilBERT base models for CoQA as extractive models, within the framework of \citet{Wolf2019HuggingFacesTS} and following \newcite{2019arXiv190311848W}, who produce the highest results with a BERT-based extractive model on CoQA.
An extractive model does not generate the answer as an abstractive model would, but selects the span in the document that best matches the gold answer. In order to train our extractive models, the substrings of the rationales which are most similar to the gold answers (as measured by F1) are selected as the training labels.

Following a standard procedure, a linear classifier head is added on top of BERT with ReLU activation which classifies every token in the input sequence as start or end of the answer span. Another linear classifier predicts whether each token in the input span falls within the rationale span or outside of it. Finally, one more classifier predicts whether the example is \textsc{freeform},
has a \textsc{yes/no} answer or is \textsc{unanswerable}. \textsc{Yes/no/unanswerable} answers are used instead of the predicted span if the model predicts the latter classes with higher confidence than the start and end tokens of the answer. Models are trained for 4 epochs (taking a few hours on a single GeForce GTX 1080 Ti GPU) with a learning rate of $3\mathrm{e}{-5}$ and AdamW optimizer~\cite{loshchilov2017decoupled}.

\section{Baseline Results and Error Analysis}
\label{sec:error}

\begin{table*}[ht]
  \centering
\begin{center}
\footnotesize{
\setlength{\tabcolsep}{4.3pt}
\renewcommand{\arraystretch}{1.1}
\begin{tabular}{l|c|cccccccccccccc}
 \toprule
\backslashbox{Model}{class}&overall&\textsc{num}&\textsc{1-5}&\textsc{neg}&\textsc{yes}&\textsc{no}&\textsc{sent}&\textsc{ant}&\textsc{ord}&\textsc{srl-}&\textsc{srl+}&\textsc{hum}&\textsc{loc}&\textsc{ent}&\textsc{surp}\\
        \midrule
 size&7983&972&128&436&790&682&2443&185&6817&3175&1242&1418&624&1089&433\\
 \midrule
 \midrule

   \textbf{RoBERTa}&81.2&\cellcolor{gray!20}76.4&\cellcolor{gray!5}40.1&\cellcolor{gray!15}72.9&\cellcolor{gray!100}89.5&\cellcolor{gray!100}85.9&\cellcolor{gray!80}83.0&\cellcolor{gray!65}82.3&\cellcolor{gray!50}81.8&\cellcolor{gray!40}80.7&\cellcolor{gray!80}83.5&\cellcolor{gray!80}83.2&\cellcolor{gray!90}83.6&\cellcolor{gray!80}83.1&\cellcolor{gray!30}78.6\\
   
   \textbf{BERT}&76.9&\cellcolor{gray!65}77.2 &\cellcolor{gray!5}41.9 &\cellcolor{gray!15}68.9&\cellcolor{gray!100}85.3&\cellcolor{gray!50} 77.3&\cellcolor{gray!80}79.0& \cellcolor{gray!30}74.8&\cellcolor{gray!50}77.2 &\cellcolor{gray!30}76.4&\cellcolor{gray!80}78.5& \cellcolor{gray!55}77.4&\cellcolor{gray!80} 80.8&\cellcolor{gray!80}79.4&\cellcolor{gray!20}74.1\\ 
   
   \textbf{DistilBERT}&66.6&\cellcolor{gray!65}69.6&\cellcolor{gray!5}36.8&\cellcolor{gray!15}56.7&\cellcolor{gray!90}82.1&\cellcolor{gray!80}71.4&\cellcolor{gray!65}69.0&\cellcolor{gray!80}71.3&\cellcolor{gray!60}67.2&\cellcolor{gray!25}65.8&\cellcolor{gray!80}70.0&\cellcolor{gray!25}64.2&\cellcolor{gray!25}65.4&\cellcolor{gray!80}70.0&\cellcolor{gray!10}58.9\\
   
   \bottomrule
 \end{tabular}
 }
 \caption{The results of the baseline models on the CoQA development set (F1 scores). The grayscale colors reflect the variation within models between \textsc{qa} classes.}
\label{tab:baseresults}
\end{center}
\end{table*}

On the development set\footnote{Due to a limitation of at most 2 submissions per week for official evaluation on CoQA, the experiments on the test set are not included.}, the RoBERTa model gets 81.2 points F1 \footnote{The results do not compare to the 89.5 F1 score of the extractive model of \citet{ju2019technical}, who do not report which RoBERTa model they use. The difference would be accounted for assuming they use RoBERTa large.}, BERT scores 76.9 F1 and falls two points short of the \citet{2019arXiv190311848W} implementation, and DistilBERT scores 66.6 F1, which establishes a baseline as this is the first work using DistilBERT on CoQA (see Table~\ref{tab:baseresults}).  

To resolve what types of linguistic inference are the hardest for the baseline models, several potentially difficult \textsc{qa} classes are analyzed. They are defined based on the findings of previous research as well as the observations of a qualitative evaluation of the errors made by the BERT model. Then, a quantitative evaluation of how the baseline models perform on each class is performed. There is ample variation in how the models score on various example classes (see Table~\ref{tab:baseresults}). Nonetheless, a noticeable trend appears of the three models failing in similar classes, with DistilBERT lagging behind BERT in most of the classes, by up to 15 points in F1 in some, and RoBERTa beating BERT by a smaller margin.

The first expected source of error for the baseline models is the inability to count listed phrases. Since the models are extractive, counting cannot fall within their capabilities.
In a rationale listing ``\textit{a poor man Ti, his son Dicky and their alien dog CJ7}", for example, the models cannot chunk the text into noun phrases and then count the chunks in order to answer `\textit{three}' to the question of how many characters there are. 
While the model performance is satisfactory on a wide range of questions with numerical answers (\textsc{num}), they fail consistently on questions with answers in the integer set between 1 and 5 (\textsc{1-5}). The \textsc{num} class is defined using a state-of-the-art rule-based question classification system from \citet{madabushi2016high}, which evaluates each \textsc{qa} based on the question alone. The contrast between the scores on the two classes can be explained by the fact that while extracting numerical answers such as dates is easy for the models, they struggle on the task of counting linguistic objects, which are usually manifested in low value integers.

The second expected problematic area is negation. The example below illustrates two ways that the model can fail in face of negation cues. 

\vspace{1.5mm}

\noindent \textbf{Rationale}: Something looked like a bird's belly [...] it was not a bird's belly [...] a bottle floated there

\noindent \textbf{Question}: What looked like a bird's belly? 

\noindent \textbf{Answer}: A bottle

\noindent \textbf{Wrong answer 1}: A bird's belly

\noindent \textbf{Wrong answer 2}: Not a bird's belly
\vspace{1.5mm}

The most general type of error is neglecting the negation cue altogether and answering the question with \textbf{Wrong answer 1}. This reflects on the model's inability to determine that the noun phrase ``\textit{a bird's belly}" falls under the scope of the negation cue ``\textit{not}". \textbf{Wrong answer 1} would be the correct answer if the phrase was not negated. The second and more rarely observed type of error reflects a lack of pragmatic knowledge as opposed to semantic or syntactic. In \textbf{Wrong answer 2} the model could be argued to have answered correctly as it is technically true that what looked like a bird's belly was not a bird's belly. However assuming Grice's maxim of quantity, which states that one should be as informative as required \cite{grice1989studies}, the answer is not satisfactory. \textbf{Wrong answer 2} is not informative at all as it has already been implied by the question. 
We define the \textsc{neg} \textsc{qa} class as containing answers that are embedded under negation.
For recognizing such answers, negation cues and their spans are detected with a BERT-based model following \citet{khandelwal2019negbert} and trained 
on the Sherlock dataset \cite{morante2012sem}. We reproduce the results on that dataset before using the model for detecting the negated spans in the background documents in the CoQA dataset to find the \textsc{neg} type answers. 
Our baseline models perform worse on the \textsc{neg} \textsc{qa} class than overall, and score much higher on questions with \textsc{yes} answers than \textsc{no} answers, which suggests that the models do not interpret negation correctly. The effect of negation on performance is particularly stark in the case of DistilBERT.

Furthermore, the \textsc{ant} class is composed of examples in which the rationale contains antonyms of the words in the question, using WordNet~\cite{wordnet}. 
Here explicit negation is not necessarily involved, however the model's ability to reason over semantic polarity is tested in this \textsc{qa} class. 
Our baseline results on class \textsc{ant} are in line with previous conclusions stating that BERT is not good at representing antonymy, as it scores lower on this class than overall. Yet interestingly, DistilBERT as well as RoBERTa perform better on this subclass of questions than overall.
We conjecture that lexical semantics is the strongest feat of BERT, therefore it is likely that DistilBERT retains most of the lexical information such as antonymy through the process of distillation. On the other hand, RoBERTa learns more about lexical features such as antonymy from the huge size of the training set. 

In addition, \textsc{sent} is a \textsc{qa} class in which the sentiment of the sentence containing the rationale is different from the sentiment of the question. The class items are determined by 
sentence splitting \cite{spacy2} and sentence-level sentiment classification \cite{Wolf2019HuggingFacesTS}.
This class is intended to capture examples where the polarity between the question and the answer can be expressed not only by negation or antonymy but also any other means, for example pragmatics. However, a qualitative analysis of the examples of the \textsc{sent} class shows that the examples which contain contradictory sentiments between the question and the answer mostly do not require one to determine the sentiment in order to answer the question correctly. For instance, the question ``\textit{How much later did he get his next job?}" has a slightly negative connotation about a long job hunt. In contrast, the rationale takes a positive outlook: ``\textit{Nearly four years later, as Obama seeks reelection, Casillas has finally landed his first full-time job, emerging out of the group known as the long-term unemployed}". The answer is ``\textit{four years}", regardless of whether that is considered too long or not. Accordingly, neither of the baseline models struggle to answer \textsc{sent} questions.

Moreover, as stated in \newcite{reddy-etal-2019-coqa}, the order of questions on the CoQA dataset follows the natural order of text, in that later questions refer, generally, to information presented towards the end of the background story. Hence, the one answering the questions ought to make inferences about what has already been discussed and where in the story they are when a given question is posed. In some cases this knowledge can be crucial for reaching the correct answer. For instance, if the story describes how ``\textit{Hans had made his way back into West Germany on foot}" and then asks whether he was in East Germany or West, one has to determine whether the question refers to the time prior to the journey or after it. In this case the answer is East Germany even though that part of the country is never mentioned in the text, which makes the example very challenging with regard to pragmatic inference. In order to evaluate how our models perform with regard to following the dialogue flow, they are evaluated on items which do in fact follow the order of the document, so that the answer to question $n$ in the text is subsequent to the answer to question $n-1$ (\textsc{ord}). It appears that all baseline models are able to infer this order to some extent and perform better on such questions than those that jump to previous passages in the text. 

\begin{figure*}[t!]
\begin{minipage}[b]{1.0\linewidth}
  \centering
  \centerline{\includegraphics[width=16cm]{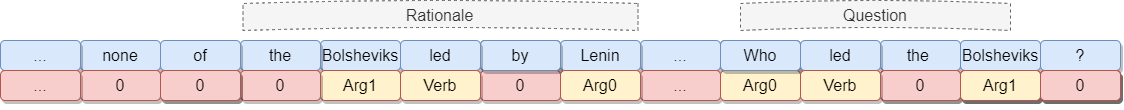}}
\end{minipage}
\caption{Illustration of the order of semantic roles in a CoQA example.}
\label{fig:flow}
\end{figure*}

Furthermore, examples are classified with regard to whether the order of the semantic roles mentioned in the question is the same (\textsc{srl+}) or different (\textsc{srl-}) to the semantic role order in the sentence containing the rationale. 
SRL is performed employing an AllenNLP~\cite{shi2019simple} model. To illustrate, Figure~\ref{fig:flow} shows an example where the roles of agent (Arg0) and patient (Arg1) are reversed in the question by means of a passive voice. 
All three models fail on such examples, scoring
lower on the \textsc{srl-} class than overall or \textsc{srl+}.
The results of the experiments show that the models find the correct answer more frequently when they can rely on the word order, avoiding the need to reason over semantic roles.

Finally, some of the observed issues are induced by the model choosing prominent entities as answers regardless of their actual relation to the question at hand. For instance, a document tells a fictional children's story wherein foods and utensils are anthropomorphised, describing how ``\textit{cereal is winning the race in a bowl of milk}", and the question is ``\textit{who is a good swimmer?}". Instead of answering with ``\textit{cereal}", the baseline BERT model chooses a human entity that is mentioned by name at the beginning of the text. 
In contrast, if ``\textit{cereal}" is substituted with a common name such as ``\textit{Mark}" in the background document, the model correctly chooses it as the answer. 
This suggests that the model relies on lexical semantics and biases about types of entities denoted by nouns more than analyze the semantic relations in the relevant sentence. 
Therefore, we define a \textsc{qa} class where the rationale contains entities that have high entropy and are thus surprising given the rest of the sentence \cite{hale2001probabilistic,levy2008expectation,smith2008optimal}, like ``\textit{cereal}" in the above example. 
In order to detect such entities, 
proper nouns (as tagged by spaCy, from \citealp{spacy2}) are masked and BERT is used to evaluate the likelihood of the original word being the filler for that mask. Words that fall below the likelihood threshold of $5\mathrm{e}{-5}$ are then deemed to be surprising entities\footnote{The threshold is selected after finetuning the model augmented with surprising word substitution, as discussed in Section~\ref{sec:enhancement}.}. All three models perform worse on questions about surprising entities (\textsc{surp}) than overall, with DistilBERT exhibiting the largest margin. 

Moreover, the classes of human (\textsc{hum}), location (\textsc{loc}) or general entities (\textsc{ent}), as classified by \citet{madabushi2016high}, test the models' ability to answer questions about entity roles.
RoBERTa and BERT's performance on these classes is higher than their overall performance. On the other hand, DistilBERT fails on \textsc{hum} and \textsc{loc} entity questions more than other \textsc{qa} types. For many \textsc{hum} and \textsc{loc} questions there are multiple entities in the text that fit the entity type. Together with the results on the \textsc{surp} class, this is an indication that DistilBERT relies on entity type more than the larger models.

The baseline results on the various classes corroborate most of the results of previous research on BERT's shortcomings. Moreover, the results show that DistilBERT mostly repeats the same mistakes and often more gravely, except for some cases of lexical semantics. DistilBERT appears to lose more of BERT's already limited representations of the formal aspects of language and have stronger biases. Finally, RoBERTa also exhibits a lack of ability to perform compositional reasoning and reaches the highest scores on the more lexical \textsc{qa} types. 

\section{Model Enhancement}
\label{sec:enhancement}

\subsection{Auxiliary tasks}

The methods for defining \textsc{qa} classes are also used as sources of linguistic knowledge which are incorporated in the baseline models to enhance their performance with regard to the respective classes.
Firstly, besides the existing \textsc{freeform}, \textsc{yes/no} and \textsc{unknown}, five additional classifiers of integer answers between 1 and 5 are defined within the model, 
as it would be impossible for the models to answer counting questions extractively.
This results in the \textit{base\#} model\footnote{The \# sign next to the model name marks that the model includes a classifier for low-valued integer answers.}. Then, four additional enhanced multitask models are built in order to tackle the issues observed in the previous section. 

For every enhanced model (\textit{negation\#}, \textit{order\#}, \textit{sentiment\#}, \textit{srl\#}), the training data is tagged with annotations of the relevant linguistic information that was also used for defining the problematic classes. For \textit{negation\#}, tokens are labelled as under the scope of negation (1) or not (0); for \textit{order\#} they are labelled as occurring after the answer to question $n-1$ (1) or not (0); for \textit{sentiment\#} they are labelled as part of a sentence with a negative sentiment (1) or not (0); while for \textit{srl\#} a multi-label setup is used where every token is labelled as either taking a particular semantic role (1) or not (0). Each of these sets of labels are then used as an additional training goal for the model.
The loss of a given additional goal is added to the main loss. 

In addition to the multitask approach, other architectures were explored for adjoining the information from the four knowledge sources. These approaches include supplying the information as an additional input feature added or concatenated to the BERT model at the level of BERT inputs themselves or the BERT model outputs. However, experiments with the latter methods showed no considerable increase in the model performance. 
Thus, the multitask approach was finally adopted for enhancing models. The multitask approach is also beneficial as the model can be applied to other test sets without the overhead of extracting the linguistic knowledge from the new set.

\subsection{Surprising Word Substitution}

One more enhanced model is produced by augmenting the training data by means of surprising word substitution (\textit{surprisal\#}). Supplementary data samples are produced by substituting surprising entities in the CoQA training set with entities that would be very likely to take their place, according to BERT. In order to ensure that the sentence structure is not affected and an entity is substituted with another entity, the substituting word was only selected if the new word was also tagged as a proper noun in the newly produced sentence. 
This procedure leads to 5880 additional samples for training. The reason behind adding these items on top of the training set instead of substituting the surprising examples is the intention to provide rare entities with better context instead of ignoring them. Many models in NLP suffer from strong social biases 
and therefore this approach attempts to level the playing field for rare entities by introducing them in the same contexts as the common entities. Such a method could potentially also be applied to larger datasets and the more early stages such as pretraining. 

Finally, the enhanced models are combined into an \textit{ensemble} in order to combine the strongest points of each model. In order to use the specialized knowledge from each model where it is relevant, the ensemble is created by selecting the model with the highest confidence for each prediction.

\section{Enhanced Model Results}
\label{sec:results}

\begin{table*}[ht]
  \centering
\begin{center}
\footnotesize{
\tabcolsep=0.11cm
\renewcommand{\arraystretch}{1.03}
\begin{tabular}{ p{1.6cm}|p{1cm}|p{0.7cm}p{0.7cm}p{0.7cm}p{0.7cm}p{0.7cm}p{0.7cm}p{0.7cm}p{0.7cm}p{0.7cm}p{0.7cm}p{0.7cm}p{0.7cm}p{0.7cm}p{0.7cm}}
 \toprule
class&overall&\textsc{num}&\textsc{1-5}&\textsc{neg}&\textsc{yes}&\textsc{no}&\textsc{sent}&\textsc{ant}&\textsc{ord}&\textsc{srl-}&\textsc{srl+}&\textsc{hum}&\textsc{loc}&\textsc{ent}&\textsc{surp}\\
  \midrule
    model&\multicolumn{15}{c}{RoBERTa}\\
    \midrule
   \textit{base}&\cellcolor{gray!40}81.2&\cellcolor{gray!40}76.4&\cellcolor{gray!40}40.1&72.9&89.5&85.9&83.0&82.3&81.8&80.7&83.5&83.2&83.6&83.1&78.6\\
    \hline
   \textit{base\#}&\cellcolor{blue!20}82.1&\cellcolor{blue!40}81.4&\cellcolor{blue!60}68.8&\cellcolor{gray!40}73.5&\cellcolor{gray!40}89.6&\cellcolor{gray!40}86.3&\cellcolor{gray!40}83.8&\cellcolor{gray!40}82.4&\cellcolor{gray!40}82.5&\cellcolor{gray!40}81.6&\cellcolor{gray!40}83.9&\cellcolor{gray!40}83.0&\cellcolor{gray!40}85.0&\cellcolor{gray!40}84.2&\cellcolor{gray!40}80.5\\
   \textit{negation\#}&\cellcolor{blue!20}81.7&\cellcolor{blue!40}81.2&\cellcolor{blue!60}69.4&\cellcolor{red!20}73.0&\cellcolor{red!20}89.5&\cellcolor{red!40}85.0&\cellcolor{blue!20}83.9&\cellcolor{blue!40}86.0&\cellcolor{red!20}82.2&\cellcolor{red!20}81.0&\cellcolor{red!20}83.2&\cellcolor{red!20}82.4&\cellcolor{red!40}83.1&\cellcolor{red!40}83.2&\cellcolor{red!20}79.8\\
  \textit{sentiment\#}&\cellcolor{blue!20}81.7&\cellcolor{blue!40}81.2&\cellcolor{blue!60}71.5&\cellcolor{red!40}71.8&\cellcolor{blue!20}90.3&\cellcolor{red!40}84.1&\cellcolor{red!20}83.6&\cellcolor{blue!40}83.5&\cellcolor{red!20}82.1&\cellcolor{red!20}81.3&\cellcolor{red!20}83.2&\cellcolor{red!20}82.1&\cellcolor{blue!20}85.4&\cellcolor{blue!20}83.2&\cellcolor{blue!40}80.7\\
   \textit{order\#}&\cellcolor{blue!40}82.3&\cellcolor{blue!40}82.7&\cellcolor{blue!60}76.0&\cellcolor{blue!20}74.3&\cellcolor{blue!20}90.0&\cellcolor{red!40}85.1&\cellcolor{blue!20}84.2&\cellcolor{blue!40}85.8&\cellcolor{blue!40}82.8&\cellcolor{blue!20}82.3&\cellcolor{blue!20}84.4&\cellcolor{blue!20}83.6&\cellcolor{blue!20}85.4&\cellcolor{red!20}83.8&\cellcolor{red!20}80.3\\
    \textit{srl\#}&\cellcolor{blue!40}82.2&\cellcolor{blue!40}82.6&\cellcolor{blue!60}73.3&\cellcolor{red!20}72.9&\cellcolor{red!40}88.6&\cellcolor{red!20}85.7&\cellcolor{blue!40}84.5&\cellcolor{blue!40}83.5&\cellcolor{blue!20}82.6&81.6&\cellcolor{red!20}83.6&\cellcolor{blue!20}83.4&\cellcolor{blue!20}85.2&\cellcolor{red!20}83.6&\cellcolor{red!20}80.4\\
     \textit{surprisal\#}&\cellcolor{blue!20}82.1&\cellcolor{blue!40}81.7&\cellcolor{blue!60}73.2&\cellcolor{red!20}72.7&\cellcolor{red!20}89.3&\cellcolor{red!20}86.0&\cellcolor{red!20}83.6&\cellcolor{blue!40}84.5&\cellcolor{red!20}82.4&\cellcolor{blue!20}81.8&83.9&\cellcolor{red!20}82.3&\cellcolor{blue!20}84.2&\cellcolor{red!20}83.7&\cellcolor{red!40}79.4\\
    \textit{ensemble}&\cellcolor{blue!40}83.9&\cellcolor{blue!40}82.7&\cellcolor{blue!60}82.2&\cellcolor{blue!50}81.7&\cellcolor{red!50}81.7&\cellcolor{red!40}82.1&\cellcolor{red!20}83.0&\cellcolor{blue!20}82.9&\cellcolor{blue!20}82.6&\cellcolor{blue!20}82.1&\cellcolor{red!40}82.3&\cellcolor{red!20}82.4&\cellcolor{red!20}83.0&\cellcolor{red!40}82.6&\cellcolor{blue!40}83.3\\
 \midrule
 model&\multicolumn{15}{c}{BERT}\\
 \midrule   
 \textit{base}&\cellcolor{gray!40}76.9&\cellcolor{gray!40}77.2 &\cellcolor{gray!40}41.9 &68.9& 85.3& 77.3&79.0&74.8&77.2 &76.4&78.5 &77.4& 80.8&79.4&74.1\\ 
   \hline
  \textit{base\#}&76.9&\cellcolor{blue!40}79.6&\cellcolor{blue!60}63.5&\cellcolor{gray!40}69.7&\cellcolor{gray!40}86.3&\cellcolor{gray!40}80.2&\cellcolor{gray!40}79.5&\cellcolor{gray!40}79.6&\cellcolor{gray!40}77.4&\cellcolor{gray!40}75.7&\cellcolor{gray!40}79.6&\cellcolor{gray!40}77.1&\cellcolor{gray!40}79.6&\cellcolor{gray!40}77.9&\cellcolor{gray!40}72.6\\
   \textit{negation\#}&\cellcolor{blue!20}77.0&\cellcolor{blue!40}79.5&\cellcolor{blue!60}63.2&\cellcolor{blue!60}79.7&\cellcolor{red!40}84.5&\cellcolor{blue!20}81.0&\cellcolor{red!20}79.0&\cellcolor{blue!20}79.7&77.4&\cellcolor{blue!20}76.1&\cellcolor{red!20}79.5&\cellcolor{red!20}76.3&\cellcolor{blue!20}81.0&\cellcolor{blue!20}78.4&\cellcolor{blue!40}76.5\\
   \textit{sentiment\#}&\cellcolor{red!40}75.2&\cellcolor{red!20}76.9&\cellcolor{blue!40}45.6&\cellcolor{red!40}66.2&\cellcolor{red!40}81.4&\cellcolor{red!40}79.1&\cellcolor{red!40}76.7&\cellcolor{red!40}76.2&\cellcolor{red!40}75.6&\cellcolor{red!20}75.1&\cellcolor{red!40}76.8&\cellcolor{red!40}74.8&\cellcolor{red!20}78.9&\cellcolor{red!40}76.9&\cellcolor{red!40}71.6\\
 \textit{order\#}&\cellcolor{red!20}76.3&\cellcolor{red!20}76.6&\cellcolor{blue!40}45.2&\cellcolor{red!20}69.0&\cellcolor{red!40}83.2&\cellcolor{blue!20}81.1&\cellcolor{red!40}78.0&\cellcolor{red!40}78.3&\cellcolor{red!20}76.9&\cellcolor{blue!20}75.9&\cellcolor{blue!20}78.9&\cellcolor{blue!20}77.3&\cellcolor{blue!20}80.4&\cellcolor{blue!20}78.5&\cellcolor{blue!20}72.8\\
   \textit{srl\#}&\cellcolor{blue!20}77.2&\cellcolor{blue!40}80.8&\cellcolor{blue!60}67.0&\cellcolor{red!40}68.5&\cellcolor{red!40}84.4&\cellcolor{blue!40}81.2&79.5&\cellcolor{blue!50}86.1&\cellcolor{blue!20}77.5&\cellcolor{blue!40}78.6&\cellcolor{red!40}75.8&\cellcolor{red!40}76.0&\cellcolor{blue!20}81.5&\cellcolor{blue!40}79.2&\cellcolor{blue!40}74.6\\
   \textit{surprisal\#}&\cellcolor{red!20}76.7&\cellcolor{blue!40}80.5&\cellcolor{blue!60}70.4&\cellcolor{red!40}68.5&\cellcolor{red!40}84.1&\cellcolor{blue!40}81.6&\cellcolor{red!40}78.6&\cellcolor{blue!40}82.7&\cellcolor{red!20}77.0&\cellcolor{blue!20}75.9&\cellcolor{red!20}78.8&\cellcolor{red!20}76.2&\cellcolor{red!20}78.8&\cellcolor{red!20}77.1&\cellcolor{blue!40}73.7\\
   \textit{ensemble}&\cellcolor{blue!40}79.2&\cellcolor{blue!40}80.8&\cellcolor{blue!60}62.2&\cellcolor{blue!40}70.8&\cellcolor{red!20}86.2&\cellcolor{blue!40}82.8&\cellcolor{blue!20}81.1&\cellcolor{blue!20}81.5&\cellcolor{blue!40}79.6&\cellcolor{blue!40}78.7&\cellcolor{blue!20}81.3&\cellcolor{blue!40}79.8&\cellcolor{blue!40}82.7&\cellcolor{blue!40}80.7&\cellcolor{blue!40}76.2\\
   \midrule
    model&\multicolumn{15}{c}{DistilBERT}\\
    \midrule
   \textit{base}&\cellcolor{gray!40}66.6&\cellcolor{gray!40}69.6&\cellcolor{gray!40}36.8&56.7&82.1&71.4&69.0&71.3&67.2&65.8&70.0&64.2&65.4&70.0&58.9\\
    \hline
   \textit{base\#}&\cellcolor{blue!20}66.9&\cellcolor{red!20}69.5&\cellcolor{blue!20}37.0&\cellcolor{gray!40}58.0&\cellcolor{gray!40}82.8&\cellcolor{gray!40}72.4&\cellcolor{gray!40}69.5&\cellcolor{gray!40}70.9&\cellcolor{gray!40}67.4&\cellcolor{gray!40}65.9&\cellcolor{gray!40}69.7&\cellcolor{gray!40}65.0&\cellcolor{gray!40}66.2&\cellcolor{gray!40}70.2&\cellcolor{gray!40}60.0\\
   \textit{negation\#}&\cellcolor{red!40}65.3&\cellcolor{blue!20}70.1&\cellcolor{blue!50}45.4&\cellcolor{blue!20}58.3&\cellcolor{red!40}80.2&\cellcolor{blue!20}73.2&\cellcolor{red!40}68.4&\cellcolor{red!40}68.9&\cellcolor{red!40}66.0&\cellcolor{red!40}64.6&\cellcolor{red!20}69.6&\cellcolor{red!40}62.2&\cellcolor{red!40}63.3&\cellcolor{red!40}67.2&\cellcolor{red!40}58.7\\
  \textit{sentiment\#}&\cellcolor{red!40}64.9&\cellcolor{red!40}68.6&\cellcolor{red!40}35.4&\cellcolor{red!40}54.9&\cellcolor{red!40}80.7&72.4&\cellcolor{red!40}68.1&\cellcolor{red!40}66.8&\cellcolor{red!40}65.4&\cellcolor{red!40}64.6&\cellcolor{red!40}67.5&\cellcolor{red!40}62.0&\cellcolor{red!40}61.5&\cellcolor{red!40}67.6&\cellcolor{blue!20}60.9\\
   \textit{order\#}&\cellcolor{red!40}64.5&\cellcolor{red!20}69.3&\cellcolor{red!40}34.9&\cellcolor{red!40}53.5&\cellcolor{red!40}81.4&\cellcolor{blue!20}72.5&\cellcolor{red!40}67.1&\cellcolor{red!40}69.2&\cellcolor{red!40}65.3&\cellcolor{red!40}64.3&\cellcolor{red!40}66.8&\cellcolor{red!40}60.2&\cellcolor{red!40}62.3&\cellcolor{red!40}66.5&\cellcolor{red!20}59.9\\
    \textit{srl\#}&\cellcolor{red!40}65.8&\cellcolor{red!20}69.0&\cellcolor{blue!40}39.1&\cellcolor{red!40}56.4&\cellcolor{blue!20}83.4&\cellcolor{red!20}72.0&\cellcolor{red!40}68.4&\cellcolor{blue!40}72.2&\cellcolor{red!40}66.5&\cellcolor{red!20}65.3&\cellcolor{red!40}68.1&\cellcolor{red!40}62.9&\cellcolor{red!40}62.8&\cellcolor{red!40}68.7&\cellcolor{red!40}57.9\\
     \textit{surprisal\#}&\cellcolor{red!20}66.8&\cellcolor{blue!20}70.3&\cellcolor{red!40}35.6&\cellcolor{blue!40}59.2&\cellcolor{blue!20}82.9&\cellcolor{red!20}72.0&\cellcolor{red!20}69.0&\cellcolor{blue!40}72.0&\cellcolor{red!20}67.1&65.9&\cellcolor{red!20}69.2&\cellcolor{red!40}63.8&\cellcolor{red!20}65.8&70.2&\cellcolor{blue!20}60.4\\
    \textit{ensemble}&\cellcolor{blue!40}68.8&\cellcolor{red!40}65.7&\cellcolor{blue!60}65.8&\cellcolor{blue!40}65.3&\cellcolor{blue!45}64.9&\cellcolor{blue!40}66.8&\cellcolor{red!40}67.4&\cellcolor{red!40}66.1&\cellcolor{red!40}65.7&\cellcolor{blue!40}66.9&\cellcolor{red!40}64.5&\cellcolor{blue!40}67.2&\cellcolor{blue!40}67.8&\cellcolor{red!40}67.3&\cellcolor{blue!50}67.5\\
   \bottomrule
 \end{tabular}
 }
 \caption{The results of the baseline and enhanced models on the CoQA development set (F1 scores). The heatmap colors reflect the variation within \textsc{qa} classes between models. The results of the \textit{base\#} models should be compared to the \textit{base} results in gray to see the effect of adding the numerical answer classifier, whereas the remaining models should be compared to the \textit{base\#} results in gray in order to see the effects of the additional linguistic knowledge.}
\label{tab:enhanceresults}
\end{center}
\end{table*}

The results of all the enhanced models on all \textsc{qa} classes on the development set are presented in Table~\ref{tab:enhanceresults}. 
BERT and RoBERTa gain most in terms of F1 with the counting model on counting questions (\textit{base\#} on \textsc{1-5}), while DistilBERT only improves on these questions remarkably with the \textit{ensemble} model, requiring more auxiliary resources than the larger models for this level of abstraction. 

Moreover, BERT appears to learn formal aspects of semantics in the multitask setting. The \textit{negation\#} model improves the results on the answers that require interpreting negation (\textsc{neg} and \textsc{no}) while the \textit{srl\#} model improves on the \textsc{qa} class with semantic roles in a different order in the question and the answer (\textsc{srl-}). 
In the meantime RoBERTa does not improve on either. 
One might say that RoBERTa has learnt the abstract linguistic representations already, however its \textit{base} results show that it makes many of the same mistakes as BERT and DistilBERT on \textsc{neg} and \textsc{srl-}. In fact, BERT outperforms RoBERTa on the \textsc{neg} class when enhanced with the explicit information about negation (the \textit{negation\#} model). 
This, combined with the fact that 
RoBERTa gets a big improvement on \textsc{neg} only when the various linguistic features are combined into an \textit{ensemble},
suggests that RoBERTa mostly relies on better lexical representations for its higher scores, which is only outweighed when many compositional semantics cues are provided.
Similarly, DistilBERT only gains a small boost over the baseline on \textsc{neg} and \textsc{no} \textsc{qa} classes with \textit{negation\#} and also requires an \textit{ensemble} to improve on \textsc{SRL-}.

On the other hand, BERT and DistilBERT improve on the \textsc{hum} and \textsc{loc} classes with the \textit{ensemble} models, demonstrating an ability to improve its lexical representations. RoBERTa does not yield an improvement in this case, however even its \textit{base} model performs relatively well on these classes. 

Furthermore, BERT and DistilBERT do not get a boost in the cases of pragmatics, namely \textit{sentiment\#} and \textit{order\#}.
In contrast, RoBERTa gets a boost over the \textsc{ant} class from the \textit{sentiment\#}, and gains the largest increases across almost all classes from \textit{order\#}. It appears that RoBERTa can improve on its already high score on items containing antonymy relying on more pragmatic aspects of lexical semantics, and also is the most receptive to the pragmatic aspects of dialogue in CoQA. 

Moreover, the model trained on the dataset which was augmented through surprising word substitution (\textit{surprisal\#}) improves over \textit{base\#} on the class with surprising entities (\textsc{surp}) with BERT and DistilBERT. This shows that the method helps the models generalize better to cover new examples with surprising entities and get rid of some of the biases about entities. 
Interestingly, in the case of BERT the largest boosts in the \textsc{surp} class are produced by the \textit{negation\#} and \textit{srl\#} models, showing that focusing on compositional information such as semantic roles and negation helps the model to be less biased towards very prominent lexical information of stereotypical entities as discussed in Section~\ref{sec:error}. In contrast, in the case of RoBERTa, \textit{surprisal\#} does not yield an improvement on the \textsc{surp} class. RoBERTa requires all of the enhanced models to be combined into an \textit{ensemble} in order to get rid of the biases that all three models exhibit, suggesting that its focus on (biased) lexical representations is stronger than BERT or DistilBERT's.

Finally, the \textit{ensemble} models perform better on virtually all classes and provide a better overall score. This is to be expected as the enhanced models, while performing at a similar level, make different errors and complement each other with their respective specializations. 

\section{Conclusion}
\label{sec:conclusion}

By and large, this paper provides additional evidence that models like BERT lack linguistic abstraction abilities, often relying on superficial features such as entity name biases or word order to answer questions in CoQA. What is more, we find that while RoBERTa improves over BERT's performance, most of its gain comes from better lexical representations and it appears to fall short of solving the compositional semantic issues.
Furthermore, we provide the first evaluation and analysis of DistilBERT on CoQA, showing that DistilBERT, more so than BERT, relies on lexical information most and lacks capacity to learn compositional representations. 
In addition, we show that all three models 
can to a varying extent be enhanced by feeding them linguistic knowledge through a multitask approach. Even a small amount of training data for linguistic information such as negation can provide a very large boost to the model performance on the \textsc{qa} classes which rely on that information. 

\bibliographystyle{acl_natbib}
\bibliography{emnlp2020}

\end{document}